\documentclass[11pt]{article}
\usepackage{acl2012}
\usepackage{times}
\usepackage{latexsym}
\usepackage{amsmath}
\usepackage{multirow}
\usepackage{url}
\usepackage{graphicx}

\usepackage{enumitem}
\setlist{nolistsep}

\newcommand{\tuple}[1]{\ensuremath{\left \langle #1 \right \rangle }}

\title{Task-specific Word-Clustering for Part-of-Speech Tagging}

\author{Yoav Goldberg \\
  Google, Inc. \\
  {\tt yogo@google.com} \\
  {\tt yoav.goldberg@gmail.com} } 

\date{}

\begin{document}
\maketitle

\begin{abstract}
   While the use of cluster features became ubiquitous in core NLP tasks,
   most cluster features in NLP are based on distributional similarity.
   We propose a new type of clustering criteria, specific to the task of part-of-speech tagging.  Instead of distributional similarity, these clusters are based on the behavior of a baseline tagger when applied to a large corpus.  These cluster features provide similar gains in accuracy to those achieved by distributional-similarity derived clusters.  Using both types of cluster features together further improve tagging accuracies.  We show that the method is effective for both the in-domain and out-of-domain scenarios for English, and for French, German and Italian.  The effect is larger for out-of-domain text.
\end{abstract}

\vspace{-10pt}
\section{Introduction}

The limited amounts of annotated training data available for supervised learning call for semi-supervised learning approaches, which aim to leverage the vast amounts of readily available unannotated data in order to improve the accuracies of supervised systems.

In natural-language processing, a simple and popular method for semi-supervised learning is based on word clustering \cite{miller2004,koo2008,turian2010}: words in a large corpus are clustered into equivalence classes based on some (usually distributional) criteria, and the induced classes are then used as additional features in a supervised learning model.  The use of cluster-based features was demonstrated to improve the accuracies of many NLP tasks, including parsing \cite{koo2008,candito2009}, named-entity recognition \cite{miller2004,turian2010,lin2009,chrupala2011}, classification of semantic relations \cite{chrupala2011} and machine-translation \cite{uszkoreit2008}.

Word clusters are usually induced based on a distributional-similarity criteria: words are clustered based on the words that tend to occur before or after them.  
Clusters produced by the Brown clustering algorithm \cite{brown1992} are an example of commonly used distributional clustering features.  In this model, words are clustered by means of a probabilistic cluster-based language model.  A more scalable distributional clustering algorithm is introduced by Uszkoreit and Brants \shortcite{uszkoreit2008} who use a parallel implementation of the Exchange algorithm to cluster words based on word-to-cluster transitions.  When used as features, clusters derived using the Exchange algorithm are as effective as those derived by the Brown algorithm.  Other types of distributional clustering algorithms rely on word embeddings \cite{collobert:2008,mnih.hinton:2007}.  Turian et al. \shortcite{turian2010} find that embedding-based distributional clusters tend to underperform Brown-type clusters.

The distributional-similarity hypothesis underlying all these algorithms is that words in similar contexts behave in a similar manner.  The notion of similarity is vague and is not specific to any particular task.  In practice, distributional-similarity based clusters show a mix of semantic and syntactic properties.  Can we design word-clusters capturing properties which are relevant for a specific task?

We focus on the task of part-of-speech (POS) tagging, and present a novel \textit{task-specific} clustering criteria: words are clustered based on the behavior of a baseline tagger when applied to a large body of text.  

One of the most useful sources of information for tagging a given word $w$ with a tag $t$ is the weighted ambiguity class of the word, as represented by the conditional tagging distribution $p(T=t|w)$. 
Our first kind of clusters aim to capture exactly this information: we cluster words based on their empirical $p(T=t|w)$ distributions, as observed over a large automatically tagged corpus.

The tag of a word $w$ can be predicted to some extent also by the previous word $w_{-1}$ or the following one $w_{+1}$.  We can create word-clusters to capture these sources of information by clustering words based on the empirical distributions $p(T_{+1}=t|w)$ and $p(T_{-1}=t|w)$, and using these clusters to represent the previous and following word respectively.

The approach is related to self-training in that we use the tagger's own prediction in order to improve it.  However, in contrast to self training, we use statistics \textit{derived from} the tagger's output as \textit{additional features} for supervised training.

For POS-tagging, our task-specific clusters are as effective as those derived by a lexical distributional-similarity criteria when used on their own, and have a cumulative effect when both kinds of clusters are used together.  Moreover, the task-specific clusters serve as a very good proxy to word identity for the purpose of POS-tagging, and we can train completely unlexicalized POS-tagging models without sacrificing accuracy.

\section{Method}

Our training protocol is as follows:

\noindent 1) Train a supervised tagger on POS-annotated text.\\
   2) Use the tagger to annotate large amounts of raw \\ \hspace*{0.6em} text.\\
   3) Collect $(W,T)$ counts from the automatically \\ \hspace*{0.6em} tagged text.\\
   4) For a word $w$ occurring over $k$ times, compute: \\
   \hspace*{0.5em} $p(T=t|w) = count(w,t) / \sum_{t'}count(w,t')$\\
   5) Cluster words based on $p(T=t|w)$.\\

\noindent We then use the derived clusters as additional features in a discriminatively-trained sequence-tagger.

When clustering, we encode the conditional $p(T=t|w)$ distribution for each word $w$ as a $|T|$-dimensional vector in which the $i$th entry is the conditional probability $p(T=t_i|w)$, and cluster words based on the Euclidean distance between their vectors: 
\\ \vspace{-10pt}
\[
dist(w_1,w_2) = \sqrt{\sum_{t_i \in T}{(p(t_i|w_1) - p(t_i|w_2))^2}}
\]
\\
\noindent
We use the K-means clustering algorithm with the initialization procedure described in \cite{kmeanspp}, which stochastically favors cluster centers that are far apart from previously chosen centers.

Words provide weak signals regarding the POS-tag of the next or previous word.  
We produce clusters based on the distributions $p(T_{-1}=t|w)$ and $p(T_{+1}=t|w)$ in a similar fashion.

\section{Details and Experiments}

\vspace{-5pt}
\subsection{Parameters}

In all the experiments, we set the word frequency threshold $k$ to be 100.  Due to the large size of our unannotated corpus, we still remain with very large vocabulary sizes (see Table \ref{tbl:unannotated-corpora}).  We run the K-means algorithm for 100 iterations, and cluster the words into 256 classes. 
While the baseline-tagger features are tuned for good accuracy, we did not perform all but minimal tuning of the extended cluster features, and did not tune any of the other parameters.

\subsection{Tagger}

We use a first-order linear-chain sequence-tagger\footnote{Most previous work on POS-tagging, e.g. \cite{ratnaparkhi1996,brants.00,collins2002,toutanova.et.al.03} use at-least a second-order model for their better results.  In contrast, we use a first-order model which is much faster.
Thus, our tagging results are lower than reported in previous work evaluating on the WSJ corpus (our train/test split is also somewhat different).  Our primary interest in this work is not in demonstrating state-of-the-art tagging accuracies on the WSJ corpus but rather examining the contributions of different cluster features to the tagger accuracy on diverse corpora.}, trained using the averaged structured-MIRA algorithm.  The features include distributional clusters derived from the unannotated corpora using the Exchange algorithm and are detailed in Table \ref{tbl:tagger-features}.  Throughout the presentation, all features are assumed to be conjoined with the tag to be predicted.

\begin{table}[h]
\begin{center}
   \scalebox{0.7}{
   \begin{tabular}{l|c}
      Type & Templates \\
      \hline
      \hline
      Lexical & $w_0$ \\
      Signature & $pref(1)$ $pref(2)$ $pref(3)$ \\
      & $suf(1)$ $suf(2)$ $suf(3)$ \\
      & $capitalization$ $hyphen$ \\
      Transition & $t_{-1}$ \\
      \hline
      $\rho$ Cluster-Dist & $\rho_0$ \hskip 0.8em $\rho_{-1}$ \hskip 0.8em $\rho_{-2}$ 
      \hskip 0.8em $\tuple{\rho_{-1}, \rho}$ \hskip 0.8em $\tuple{\rho_{-2}, \rho_{-1}}$ \\
      \hskip 0.5em +Transition & $\tuple{\rho_0, t_{-1}}$ \hskip 1em $\tuple{\rho_{-1}, t_{-1}}$ \\
      \hline
   \end{tabular}}
   \caption{Tagger features for the baseline tagger. pref(n) and suf(n) are prefixes/suffixes of length $n$ of the current word $w_0$.  The distributional-similarity features $\rho$ are derived using the algorithm of \cite{uszkoreit2008}.}  
   \label{tbl:tagger-features}
\end{center}
\end{table}

\vspace{-5pt}
\subsection{Datasets}

\paragraph{Annotated data} For English, we use the following annotated corpora:

   \noindent\textbf{WSJ}  The WSJ portion of the Penn Treebank corpus \cite{marcus.etal.93} is used to train all of our English tagging models.  We train on Sections 2-21, and evaluate on Section 22.\\
   \noindent\textbf{Brown} (BRN) The entire Brown corpus portion of the Penn Treebank is used for evaluation.\\
   \noindent\textbf{Questions} (QTB) The QuestionBank \cite{qtb} contain 4,000 questions, which we use for evaluation,\\
   \noindent\textbf{Football} (FTBL) We report results on the development set (185 sentences) of the Football corpus of \cite{foster2010}.  In one experiment we use the test section (170 sentences) as additional training data. \\
   \noindent\textbf{Web} The entire \textit{web} portion of the Ontonotes corpus \cite{weischedel2011} is used for evaluation.\footnote{Note that the Ontonotes corpus is systematically different from the training corpus in several aspects, including using both the ``IN'' and ``TO'' tags for the word ``to'' depending on its usage (in the Penn Treebank all, \textit{to} is consistently tagged as TO), and the introduction of additional tags for hyphens and non-sentence-final punctuation.  While one could get vastly improved accuracies on this dataset by specifically addressing these issues, we did not do so in the current work as our primary interest is comparing the effect of the various cluster features on tagging accuracy.}

In most experiments we train our tagger on the training set of the WSJ corpus and reserve the other datasets for evaluation.  The baseline tagger is always trained on the WSJ training set.

\vspace{2pt}

\noindent\textbf{German} We use data and splits from the CoNLL 2006 shared task \cite{conll.x}.\\
\noindent\textbf{French} We use the French Treebank \cite{ftb2004} with splits defined in Candito et al. \shortcite{ftb2010}.\\
\noindent\textbf{Italian} We use data and splits from the CoNLL 2007 shared task \cite{conll07}.

\paragraph{Unannotated Data} We use one year of newswire articles from multiple sources from a news aggregation website for each language.  The datasets range in size from 19 to 0.5 billion tokens.  The unannotated data is summarized in Table \ref{tbl:unannotated-corpora}.

\begin{table}
   \center
   \scalebox{0.8}{
   \begin{tabular}{lccc}
      Language & Domain & \#Tokens & Vocabulary \\
      \hline
      \hline
      English & News &    $19 \times 10^9$   & 649K \\
      German  & News &    $2.5 \times 10^9$  & 386K \\
      French  & News &    $1.4 \times 10^9$  & 165K \\ 
      Italian & News &    $0.5 \times 10^9$  & 116K \\
   \end{tabular}}
   \caption{Details of unannotated data.  Vocabulary is the number of token-types appearing more than 100 times.}
   \label{tbl:unannotated-corpora}
\end{table}

\section{Results}

\subsection{English}

In the first set of experiments we test the effectiveness of the Task-based clustering method on both in-domain and out-of-domain English data.  

We begin by distilling the amount of information captured by the different clusters.  To this end, we train models with the simplest set of features possible: for each sequence position we consider the lexical item $w_0$, the transition feature $t_{-1}$, and zero or more cluster features.  We also train models including the cluster features but not the lexical items.  We evaluate the models on the different English datasets.  Table \ref{tbl:simple-results} detail the results.

\begin{table}[h!]
   \scalebox{0.9}{
   \begin{tabular}{l|ccccc}
      Features  & WSJ & QTB & BRN & FTBL & Web \\
      \hline
      \hline
      $t_{-1}$ $w_0$             &  94.97 & 85.93 & 91.29 & 89.79 & 88.77 \\
      \hline
      $t_{-1}$ $w_0$ $\rho_0$    & 96.01 & 88.28 & 94.11 & 91.69 & 91.13 \\
      $t_{-1}$ $w_0$ $\zeta_0$   & \textbf{96.42} & \textbf{89.74} & \textbf{94.95} & 92.85 & \textbf{92.17} \\
      $t_{-1}$ $w_0$ $\eta_{-1}$ & 95.21 & 86.07 & 91.49 & 89.58 & 89.10 \\
      $t_{-1}$ $w_0$ $\tau_{+1}$ & 95.30 & 86.34 & 91.64 & 89.73 & 89.04 \\
      \hline
      $t_{-1}$ $\rho_0$ & 94.37 & 85.66 & 92.17 & 89.49 & 88.97 \\
      $t_{-1}$ $\zeta_0$ & 96.14 & 89.24 & 94.74 & \textbf{93.17} & 92.06 \\
       \end{tabular}}
   \caption{Tagging accuracies with minimal feature sets. $\rho$:~distributional-clusters, $\zeta$:~$p(t|w)$-clusters, $\eta$:~$p(t_{+1}|w)$-clusters, $\tau$:~$p(t_{-1}|w)$-clusters.  All models are trained on the training portion of the WSJ corpus. }
   \label{tbl:simple-results}
\end{table}

Note that this is not exactly a domain-adaptation scenario, as all the unannotated data is from the Newswire domain.  Still, the cluster features contribute to tagging accuracies across all the datasets.  When the current word $w_0$ is present as feature, the distributional clusters $\rho_0$ is somewhat less informative than the task-specific clustering $\zeta_0$, which is based on $p(t|w)$.  The cluster features of the next and previous words ($\eta_{-1}$ and $\tau_{+1}$) are expectedly less informative than the cluster associated with the current word, but still contain some predictive information.  When we exclude the word from the feature set and rely only on the cluster information (the last two rows of the table), the task-specific clusters $\zeta_0$ do particularly well -- compensating almost completely over the missing word identity information.  The models relying solely on the previous tag and the task-specific cluster ($t_{-1}$ $\zeta_{0}$) are significantly better than the models relying on the previous tag and the explicit word identity ($t_{-1}$ $w_0$).

We then proceed to evaluate the effectiveness of the cluster features in the context of a richer feature set.  We use the feature-sets described in Tables \ref{tbl:tagger-features} and \ref{tbl:tagger-features2}.
We consider different subsets of the cluster features.
Results are presented in Table \ref{tbl:en-results}.

\begin{table}[h]
\begin{center}
   \scalebox{0.7}{
   \begin{tabular}{l|c}
      Type & Templates \\
      \hline
      \hline
      $\zeta$ Cluster $p(t|w)$ & $\zeta_0$ \hskip 0.8em $\zeta_{-1}$ \hskip 0.8em $\zeta_{-2}$ 
      \hskip 0.8em $\tuple{\zeta_{-1}, \zeta_0}$ \hskip 0.8em $\tuple{\zeta_{-2}, \zeta_{-1}}$ \\
      +Transition & $\tuple{\zeta_0, t_{-1}}$ \hskip 1em $\tuple{\zeta_{-1}, t_{-1}}$ \\
      \hline
      \hline
      $\eta$ Cluster $p(t_{+1}|w)$ & $\eta_{-1}$  \\
      \hskip 0.5em +Transition & $\tuple{\eta_{-1}, t_{-1}}$ \\
      \hline
      $\tau$ Cluster $p(t_{-1}|w)$ & $\tau_{+1}$ \\
      \hline
   \end{tabular}}
   \caption{Additional cluster features.}
   \label{tbl:tagger-features2}
\end{center}
\end{table}

\begin{table}[h]
   \scalebox{0.77}{
   \begin{tabular}{l|cccccc}
            & No & Dist & Task & Dist+Task & All & All \\
            & clusters            & $\rho$ & $\zeta$ & $\rho$ $\zeta$ & $\rho$ $\zeta$ $\eta$ $\tau$ & (no $w_0$) \\
      \hline
      \hline
      WSJ       & 96.35 & 96.90 & 96.82 & \textbf{97.01} & \textbf{97.02} & \textbf{97.02} \\
      QTB & 88.86 & 90.74 & 90.50 & 90.83 & \textbf{90.93} & \textbf{90.93} \\
      BRN     & 94.37 & 95.57 & 95.48 & 95.68 & \textbf{95.72} & \textbf{95.70} \\
      FTBL  & 91.96 & 93.38 & 93.44 & 93.74 & 93.80 & \textbf{94.03} \\
      Web       & 91.38 & 92.81 & 92.82 & 92.99 & \textbf{93.05} & \textbf{93.05} \\
   \end{tabular}}
   \caption{Tagging accuracies using the different clusterings within a rich feature set.  All models are trained on the training portion of the WSJ corpus. Last column contain all the features but the lexical one.}
   \label{tbl:en-results}
\end{table}

Expectedly, using the richer feature-sets improve results for all models. The cluster features still contribute to tagging accuracies across all the datasets.  The contribution of the task-based clusters (Task) is similar but a bit lower than that of the distributional clusters (Dist), but results improve when the two clustering approaches are combined (Dist+Task).  Adding the task based clustering of neighboring words (All) further improve the results on most datasets.  The largest improvements are observed on the out-of-domain datasets.  Somewhat surprisingly, dropping the explicit lexical feature $w_0$ (last column) does not hurt performance, and even significantly improve it on the Football dataset.

\vspace{-5pt}
\subsection{English -- Additional Training Data}

In the next experiment, we target the situation in which we have a small amount of annotated data in an interest-domain in addition to the larger amount of out-of-domain data. We use the test-set of the Football dataset as additional in-domain training material.
Results are presented in Table \ref{tbl:adapt}. As expected, using the additional in-domain training data improve the results.  However, the contribution of the additional data is small, as most of the gap is already covered by the cluster features.  When using all the cluster features but no lexicalization (last column) training on WSJ alone outperform the joint training.

\begin{table}[h]
   \scalebox{0.77}{
\begin{tabular}{l|cccccc}
Train          &  No & Dist & Task & Dist+Task & All & All \\
            &     clusters        & $\rho$ & $\zeta$ & $\rho$ $\zeta$ & $\rho$ $\zeta$ $\eta$ $\tau$ & (no $w_0$) \\
\hline
\hline
WSJ            & 91.96        & 93.38 & 93.44 & 93.74 & 93.80 & \textbf{94.03} \\
+FTBL   & 92.31        & 93.47 & 93.50 & 93.92 & \textbf{93.94} & 93.83 \\
\end{tabular}}
\caption{Adaptation results to the Football domain when training on both datasets.}
\vspace{-10pt}
\label{tbl:adapt}
\end{table}

\vspace{-5pt}
\subsection{German, French and Italian}

We observe similar trends on languages other than English (Table \ref{tbl:multilingual}). 
The additional task-specific cluster features improve performance across all languages.

\begin{table}[h]
   \scalebox{0.8}{
\begin{tabular}{l|ccccc}
Language       & No Clusters & Dist & Task & Dist+Task & All \\
            &             & $\rho$ & $\zeta$ & $\rho$ $\zeta$ & $\rho$ $\zeta$ $\eta$ $\tau$ \\
\hline
\hline
German         & 96.48       & 97.68 & 97.70 & 97.84 & \textbf{98.00} \\
French         &   96.45    & 97.55 & 97.54 & 97.66 & \textbf{97.74} \\
Italian        & 93.58       & 96.00 & 96.15 & \textbf{96.43} & 96.39 \\
\end{tabular}}
\caption{Tagging results for German, French and Italian.}
\vspace{-10pt}
\label{tbl:multilingual}
\end{table}

\vspace{-5pt}
\section{Conclusions}

We presented a task-specific word clustering method for POS-tagging.  The method is effective across domains and languages.  The automatically derived clusters capture the essence of the lexical items with respect to the task to the extent that the cluster features can replace the actual lexical items.  We would like to see task-specific clusterings for other, more challenging tasks.

\bibliographystyle{acl2012}
\bibliography{bib}
\end{document}